\documentclass[10pt,twocolumn,letterpaper]{article}

\usepackage[pagenumbers]{cvpr} 

\usepackage{times}
\usepackage{epsfig}
\usepackage{graphicx}
\usepackage{amsmath}
\usepackage{amssymb}
\usepackage{booktabs}
\usepackage{multirow}
\usepackage{animate}
\usepackage{float}
\usepackage{etoolbox}
\usepackage{array}
\usepackage{textcomp}
\usepackage{stfloats}
\usepackage{url}
\usepackage{verbatim}
\usepackage{xcolor}

\definecolor{cvprblue}{rgb}{0.21,0.49,0.74}
\definecolor{cityblue}{RGB}{128, 159, 225}
\definecolor{citypink}{RGB}{227, 108, 194}

\newcommand{\coloredref}[2]{\hyperref[#2]{\textcolor{#1}{\ref*{#2}}}}
\newcommand{\coloredcite}[2]{\hyperref[#2]{\textcolor{#1}{\cite{#2}}}}

\usepackage{newfloat}
\usepackage{listings}
\usepackage{cuted} 
\usepackage{capt-of}

\usepackage{algorithm}
\usepackage{algorithmic}

\DeclareCaptionStyle{ruled}{labelfont=normalfont,labelsep=colon,strut=off}
\floatstyle{ruled}
\newfloat{listing}{tb}{lst}{}
\floatname{listing}{Listing}

\def\eg{\emph{e.g.}}
\def\ie{\emph{i.e.}}

\def\name{StyleMe3D}

\usepackage[pagebackref,breaklinks,colorlinks,citecolor=cvprblue]{hyperref}

\newcommand\blfootnote[1]{%
  \begingroup
  \renewcommand\thefootnote{}\footnote{#1}%
  \addtocounter{footnote}{-1}%
  \endgroup
}

\begin{document}

\title{\name{}: Stylization with Disentangled Priors by Multiple
Encoders on 3D Gaussians}

\author{
Cailin Zhuang\textsuperscript{1,2,3} \quad
Yaoqi Hu\textsuperscript{3} \quad
Xuanyang Zhang\textsuperscript{2$\dagger$} \quad
Wei Cheng\textsuperscript{2} \\ 
Jiacheng Bao\textsuperscript{1} \quad 
Shengqi Liu\textsuperscript{2} \quad
Yiying Yang\textsuperscript{2} \quad 
Xianfang Zeng\textsuperscript{2} \quad
Gang Yu\textsuperscript{2} \\ 
Ming Li\textsuperscript{4$\ddagger$}
\vspace{0.3em} \\
\textsuperscript{1}ShanghaiTech University \quad
\textsuperscript{2}StepFun \quad
\textsuperscript{3}AIGC Research \quad
\textsuperscript{4}Guangming Laboratory \quad
\vspace{0.3em} \\
{\tt \url{https://styleme3d.github.io/}}
}

\maketitle

\begin{strip}
  \vspace{-1cm} 
  \centering
  \includegraphics[width=\textwidth]{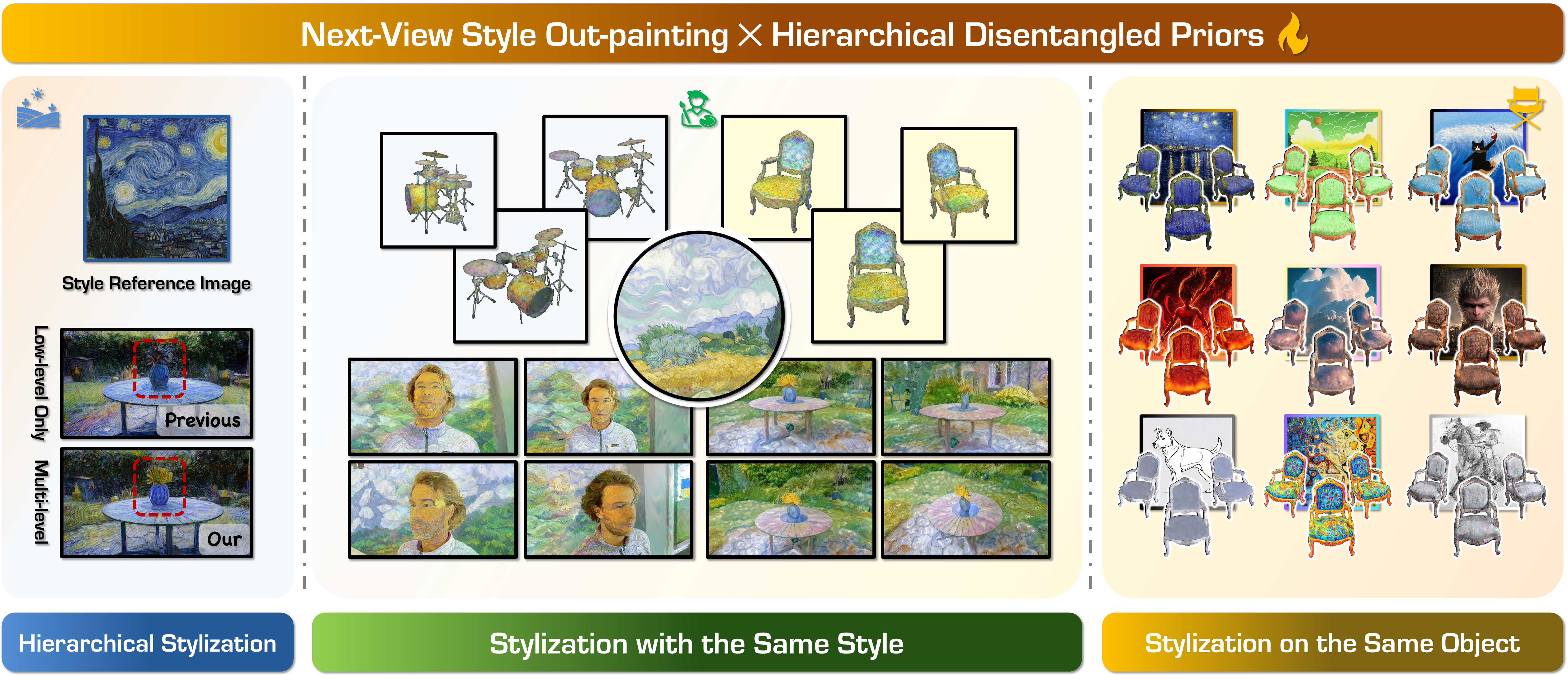}
  \captionof{figure}{Our \name{} enables hierarchical, versatile and high-quality 3D stylization across diverse styles.}
  \label{fig:banner}
\end{strip}

\blfootnote{Project lead: $^{\dagger}$X. Zhang; Corresponding author: $^{\ddagger}$M. Li.}

\begin{abstract}

Current 3D Gaussian Splatting stylization approaches are limited in their ability to represent diverse artistic styles, frequently defaulting to low-level texture replacement or yielding semantically inconsistent outputs.
In this paper, we introduce \name{}, a novel hierarchical framework that achieves comprehensive, high-fidelity stylization by disentangling multi-level style representations while preserving geometric fidelity.
The cornerstone of \name{} is Dynamic Style Score Distillation (DSSD), which harnesses latent priors from a style-aware diffusion model to provide high-level semantic guidance, ensuring robust and expressive style transfer.
To further refine this distillation process, we propose a multi-modal alignment strategy using the CLIP latent space: a CLIP-based style stream evaluator (Contrastive Style Descriptor) that enforces middle-level stylistic similarity, and a CLIP-based content stream evaluator (3D Gaussian Quality Assessment) that acts as a global regularizer to mitigate typical GS quality degradation.
Finally, a VGG-based Simultaneously Optimized Scale module is integrated to refine fine-grained texture details at the low-level.
Extensive experiments demonstrate that our method consistently preserves intricate geometric details and achieves coherent stylistic effects across entire scenes, significantly surpassing state-of-the-art baselines in both qualitative and quantitative evaluations. 

\end{abstract}

\section{Introduction} \label{sec:intro}

The advent of 3D Gaussian Splatting (GS)~\cite{kerbl20233d} has revolutionized 3D scene representation, delivering high reconstruction fidelity and real-time rendering. This advancement has broadened the horizons of 3D stylization, enabling direct scene editing towards given styles. Nevertheless, current methods remain notably limited in their capacity for expressive artistic control. They can be mainly divided into two categories, \ie, single-template color matching methods and texture transfer counterparts. A typical method of the former is Ref-NPR~\cite{zhang2023ref}, which depends solely on global style exemplars and consequently fails to capture fine-grained textures. The texture transfer methods ~\cite{galerne2024sgsst,de2021maximum,liu2024stylegaussian,zhang2024stylizedgs} adapt basic color and patterns using VGG-based representations~\cite{simonyan2014very} but lack semantic understanding, resulting in an inability to preserve coherent semantics. These shortcomings are particularly pronounced with abstract styles and complex scene layouts, as existing methods focus exclusively on distilling a single level of style representation.

To address these limitations, we introduce a novel hierarchical framework that establishes a new paradigm for semantic-aware GS stylization. Our central innovation is the integration of multi-level style understanding and transfer within a single, unified framework. By disentangling stylistic features across low, middle, and high semantic levels, our framework achieves expressive artistic transformation while strictly upholding the structural integrity of the underlying 3D content.

At the heart of our framework lies a suite of specialized modules, each tailored to address different facets of 3D stylization. For high-level representation, we leverage the latent space of a style-enhanced diffusion model~\cite{rombach2022high} to achieve robust, content-aware alignment between style exemplars and 3D geometry, marking the first application of such semantic priors in GS stylization. 
To complement this core engine, we introduce a CLIP-based dual-stream alignment mechanism that operates on the multi-modal latent space: (1) a Style Stream, implemented via the Contrastive Style Descriptor (CSD)~\cite{somepalli2024CSD}, which is pre-trained on diverse style datasets and employs a CLIP-based evaluator to enforce mid-level stylistic consistency; and (2) a Content Stream, realized through 3D Gaussian Quality Assessment (3DG-QA)~\cite{wang2023exploring}, which leverages contrastive aesthetic priors and acts as a semantic regularizer to mitigate geometric degradation. Finally, a Simultaneously Optimized Scale (SOS)~\cite{galerne2024sgsst} module is incorporated to refine low-level texture details.
Collectively, these innovations establish a comprehensive and flexible solution that substantially advances the expressive power and quality of GS stylization.


We validate our method on both 3D object dataset (\ie, NeRF synthetic~\cite{mildenhall2021nerf}) and scene datasets (\ie, tandt db~\cite{kerbl20233d} and mip-NeRF 360~\cite{barron2022mipnerf360}), demonstrating its generalization ability across various geometric complexities and artistic styles. Our approach consistently achieves superior stylization fidelity compared to state-of-the-art (SOTA) baselines. Qualitatively, our method excels at preserving fine geometric structures while establishing scene-level stylistic coherence throughout diverse content. Quantitatively, our approach achieves remarkable improvements in CLIP-based style similarity and LPIPS/PSNR/SSIM metrics over prior methods.


Our main contributions are summarized as follows: \begin{itemize} 
\item \textbf{Hierarchical 3D Stylization}: We present the first multi-level prior-disentangled 3D Gaussian Splatting framework, enabling geometry-preserving stylization that captures low-level, mid-level, and high-level artistic attributes. 
\item \textbf{Semantic-Aware Style Transfer}: By leveraging a style-enhanced diffusion latent space prior (DSSD), our approach enables robust semantic alignment and expressive style transfer, surpassing the limitations of conventional VGG-based methods. 
\item \textbf{CLIP-based Dual-stream Alignment}: We introduce a aesthetic-driven synergistic alignment strategy comprising a CLIP-based Style Stream (CSD) for attribute consistency and a CLIP-based Content Stream (3DGQA) for maintaining structural and semantic fidelity. 
\end{itemize}


\begin{figure*}[t]
  \centering
    \includegraphics[width=\linewidth]{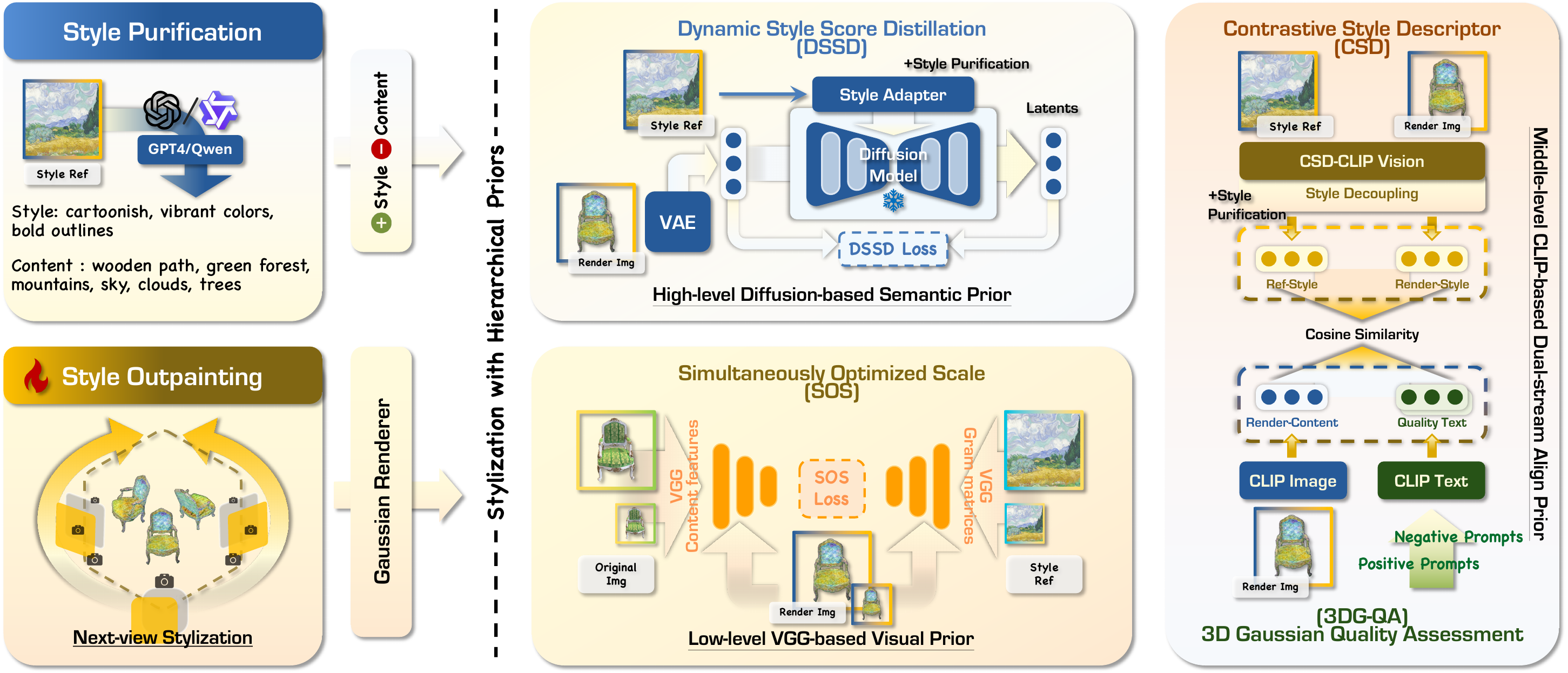}
  \caption{\small
\textbf{Overview of our \name{} framework}.
(a) \textbf{Style Purification}: Extracts and refines stylistic representations within the CLIP latent space by decoupling style from the content of the reference image.
(b) \textbf{Hierarchical Stylization}: The \textbf{Dynamic Style Score Distillation (DSSD)} module employs dynamic noise scheduling and adaptive style guidance, integrating latent losses to achieve consistent stylization step by step. Integrates three specialized components within DSSD:
\textbf{Simultaneously Optimized Scale (SOS)}: Refines low-level texture details through adaptive noise-scale optimization.
\textbf{Contrastive Style Descriptor (CSD)}: Serves as the \textit{CLIP-based Style Stream} to separates mid-level style and content via contrastive learning for style similarity score.
\textbf{3D Gaussian Quality Assessment (3DG-QA)}: Acts as the \textit{CLIP-based Content Stream} for quality-guided refinement, utilizing antonymic semantic prompts to preserve structural integrity. 
(c) \textbf{Progressive Next-view Optimization (Style Outpainting)}: Progressive outpainting achieves novel view style propagation. Ensures global coherent through iterative latent alignment, eliminating multi-view dependencies.
  }
  \label{fig:overview_pipeline}
\end{figure*}

\section{Related Works}
\label{sec:related}

\subsection{2D Generation and Stylization}

2D generation has rapidly advanced across generative modeling, customization, conditional control, editing, and stylization.
Initial breakthroughs in 2D synthesis with VAEs and GANs \cite{goodfellow2014generative,brock2018large,karras2019style} were furthered by diffusion models \cite{li2024playground,rombach2022high,xing2024svgdreamer,zeng2024jedi}, enhancing image quality and diversity for complex manipulation.
Stylization advances emphasize style-content separation, with cross-attention-based transfer \cite{ye2023ip,wang2024instantstyle,chung2024style} and shared attention mechanisms for coherence \cite{wu2021stylealign}. Frequency-domain techniques aid diffusion control \cite{gao2024frequency}, while Aligning style with textual cues \cite{li2024styletokenizer}, cross-domain fusion \cite{pham2024tale} and FFT-based transfer \cite{he2024freestyle} expand style applications. 
In this paper, we aim to style 3D GS and these 2D methods give us a lot of insights and priors that can be reused in the 3D field.

\subsection{3D Generation and Stylization}

Text-guided 3D generation has also advanced with Score Distillation Sampling (SDS) \cite{poole2022dreamfusion} and its variants \cite{wang2024prolificdreamer, wang2023score}, enabling controllable, diverse 3D synthesis. These techniques support artistic scene generation \cite{li2024art3d} and multimodal inputs (text and image) \cite{wang2024phidias, yan2024dreamdissector, tang2023dreamgaussian, tang2025lgm}. Recent improvements in latent diffusion models further enhance the expressiveness and creative potential of text-to-3D generation \cite{yan2024dreamdissector, zhou2024diffgs}, and more and more multi-view~\cite{cheng2022generalizable,cheng2023dna,pan2023renderme} and 3D dataset~\cite{objaverseXL,wu2023omniobject3d} still stimulate the development of this field. 


For 3D stylization, methods like \cite{huang2021learning, mu20223d} embed styles directly into 3D structures, while radiance field-based methods \cite{nguyen2022snerf, wang2023nerf, zhang2022arf} achieve style transfer through optimization for enhanced scene realism. Though HyperNet \cite{chiang2022stylizing} enables arbitrary style embedding in MLPs, it suffers from slow rendering and detail loss, while StyleRF \cite{liu2023stylerf} offers zero-shot stylization by transforming radiance field features but lacks adaptability and control.

Recent advances in 3D stylization have explored various techniques to embed artistic styles into 3D content, with reference-based methods like \cite{zhang2023ref, mei2024reference} for controlled stylization and arbitrary reference techniques \cite{zhang2024stylizedgs, liu2024stylegaussian} for flexible style transfer. Scalable 3D style transfer brings the 3d stylized resolution up to 4K by SOS Loss \cite{galerne2024sgsst}. Stylized Score Distillation \cite{kompanowski2024dream} and 3D-aware diffusion models \cite{wang2024phidias} further expand these capabilities. 

Unlike previous works, we conduct a systematic analysis of 3D GS stylization and propose a comprehensive framework to achieve hierarchical, multi-granular style transfer.

\section{Method: \name{}} 
\label{sec:method}
In this section, we elaborate on our comprehensive algorithmic framework for 3D style transfer using 2D priors. We first formally define our core task: performing style transfer on reconstructed 3D Gaussian Splatting (3D GS) representations while preserving structural fidelity in Sec.\ref{sec:task_def}. To address the inherent challenges in cross-dimensional style adaptation, we propose \name{}, a systematic framework comprising mixture of four encoders that collectively resolve critical challenges in 3D style migration from Sec.\ref{sec_dssd} to Sec.\ref{sec_qa} and is unified in Sec.\ref{sec_moe}.

\subsection{Preliminaries}
\label{sec:task_def}
We define initial 3D GS Reconstruction as a pre-trained task, while redefining 3D GS stylization as a post-training task. Unlike conventional 3D generation tasks that begin from scratch, our approach applies stylization to pre-reconstructed 3D gausion for both 3D objects and scenes, allowing for enhanced control over style application while preserving the underlying geometry.\\
Firstly, we define the 3D GS reconstruction process as:
\begin{equation}
    \mathop{\min}_{\Theta} \frac{1}{N} \sum_{v=1}^{N} MSE(\mathcal{R}(\mathcal{C}_{v};\Theta), I_{v}^{gt})
    \label{eq_1}
\end{equation}

where $\Theta=\{(u_{i},\Sigma_{i},\alpha_{i}, c_{i,0},(c_{i,j,k})_{j,k}))\}_{i=1}^{N^{Gaussians}}$ represents the 3D gaussian, $c_{i,0}$ is the main color and $c_{i,j,k}$ is the coefficient. $\mathcal{R}(\mathcal{C}_{v};\Theta)$ means render 3D gaussian and $I_{i}^{gt}$ means the ground truth image from the viewpoint $\mathcal{C}_{v}$ respectively.

After obtaining the optimized 3D gaussian, we further formulate the 3D gaussian style transfer process with 2D prior as follows:
\begin{equation}
    \mathop{\min}_{\Theta} \frac{1}{N} \sum_{v=1}^{N} \mathcal{L}(\mathcal{R}(\mathcal{C}_{v};\Theta);\phi, R)
\end{equation}
where $\phi$ means the 2D prior and $R$ means the reference prompt, like text prompts or image prompts. $\mathcal{L}$ means the loss function to further optimize the 3D gaussian which is initialized with $\Theta$ from Eq.\ref{eq_1}.

In the style transfer task, we aim to only change the 3D gaussian stylization rather than the geometry content. We achieve geometry-style decoupling in 3D gaussian by leveraging the inherent separation of geometric and color parameters in its parametric representation. Specifically, our style transfer framework exclusively optimizes the color parameters $\Theta_{color}$ while maintaining frozen geometric attributes during the stylization process as:
\begin{equation}
    \mathop{\min}_{\Theta_{color}} \frac{1}{N} \sum_{v=1}^{N} \mathcal{L}(\mathcal{R}(\mathcal{C}_{v};\Theta);\phi, R)
\end{equation}

We further discuss how to instantiate the $\mathcal{L}$, $\phi$ and $R$ with different formulations and jointly improve the stylization effectiveness in the following sections.

\subsection{Next-view Stylization: Progressive Style Outpainting}
Progressive Style Outpainting~(PSO) is a novel style guidance method for consistent and detailed style propagation in multi-view 3D stylization~(see Fig. \ref{fig:overview_pipeline}). Using 2D style priors provided by an image stylization diffusion model~\cite{gao2024styleshot}, we redefine multi-view guidance as a progressive outpainting task. By integrating sparse-view RGB loss with dense-view SDS loss, PSO ensures consistent 3D stylization across views. Instead of random view selection, our method incrementally propagates style information to adjacent views, enhancing style coherence with each step. Specifically, PSO consists of two primary guidance modes, namely gobal guidance and local guidance. 

\noindent\textbf{Global Guidance.}~In the global guidance stage, a uniform noise level is applied to all views before stepwise reduction, defined as:
\begin{equation}
    \alpha_{\text{step}} = \frac{\left( \left\lfloor \frac{i_{\text{step}}}{n_{\text{view}}} \right\rfloor \mod n_{\text{opt}} \right)}{n_{\text{opt}}} 
\end{equation}
where \( n_{\text{view}} \) represents the total number of rendering views and \( n_{\text{opt}} \) denotes the required optimizations per view, managed iteratively by \( i_{\text{step}} \).

\noindent\textbf{Local Guidance.}~Local guidance focuses on single-view optimization, maximizing stylization quality for individual views, albeit at the potential expense of global consistency. The local guidance schedule is defined as:
\begin{equation}
    \alpha_{\text{step}} = \frac{i_{\text{step}} \mod n_{\text{opt}}}{n_{\text{opt}}}
\end{equation}

The effectiveness of these modes in balancing stylization strength and consistency is discussed in Sec.~\ref{subsec:ablation}. To maximize the stylization outcome, we combine both guidance modes for complementary strengths. 

\subsection{High-level Guidance: Dynamic Style Score Distillation} 
\label{sec_dssd}
In this section, we distill the prior from the 2D stable diffusion model~\cite{rombach2022high} and use both text and image prompt for style transfer.

\noindent\textbf{Style Purification.}
Inspired by InstantStyle~\cite{wang2024instantstyle}, we use a pre-trained CLIP model for Style Purification to isolate pure style information. In CLIP space, we filter out style-irrelevant details by subtracting content descriptors from style embeddings. Specifically, descriptions of the style reference image are generated using a captioning model (e.g., GPT-4V) to distinguish content-related descriptors. The CLIP Text Encoder extracts a \textit{Content Text Embedding} (or both content and style) from these descriptors, while the CLIP Image Encoder produces a \textit{Style Image Embedding}. Subtracting \textit{Content Text Embedding} from \textit{Style Image Embedding} (and adding \textit{Style Text Embedding}) yields a \textit{Final Style Embedding} containing only style-related information. The style purification process is shown in Fig. \ref{fig:overview_pipeline}.

\noindent\textbf{Fine Timestep Sampling.}
Fine timestep sampling enhances temporal resolution by focusing on low-noise intervals for more granular optimization, with noise progressively decreasing from high to low levels. This sampling strategy is formulated as:
\begin{equation}
    t = Round(( 1 - \alpha_{\text{step}}^{0.5} ) \cdot \text{T} ). \text{clip}(T_{\text{min}}, T_{\text{max}})
\end{equation}
where \( \text{T} \) denotes the total timesteps, with \( T_{\text{min}} \) and \( T_{\text{max}} \) setting the bounds. Higher noise initialization effectively eliminates outlier gaussian, refining the stylization outcome.

\noindent\textbf{Style Distillation.}
As shown in Fig. \ref{fig:overview_pipeline}(b). DSSD further extends score distillation by applying a dynamic CFG (Classifier-Free Guidance) \cite{ho2022classifier} scale coefficient to optimize the intensity of style guidance. Fixed CFG values can lead to oversmoothing (low CFG) or oversaturation (high CFG). To counter this, we introduce a dynamic guidance coefficient that adaptively balances fixed CFG values throughout optimization. The adaptive coefficient is defined as:
\begin{equation}
    \Delta \lambda = \max \left( 7.5, \, \lambda_{\text{max}} \cdot \left( \alpha_{\text{step}}^{2} \right) \right)
\end{equation}
With this method, we extend the SSD proposed by \cite{kompanowski2024dream}, and define the style loss in latent space as:
\begin{equation}
    \text{DSSD}_{\text{2D}}^{z} = (1 - \Delta \lambda_{s}) \epsilon_{\phi \text{2D}}(z_{t_s} | y, t_s) + \Delta \lambda_{s} \hat{\epsilon}_{\phi \text{2D}}(z_{t_s} | y, s, t_s) - \epsilon_s,
\end{equation}
where \( \epsilon_{\phi 2D}(\cdot) \) is the predicted noise by the style-based 2D diffusion prior \( \phi \).

Our Dynamic Style Score Distillation (DSSD) objective function as follows:
\begin{multline}
    \nabla_{\Theta_{color}} \mathcal{L}_{\text{DSSD}}(x=\mathcal{R}(\mathcal{C}_{v};\Theta_{color});\phi, R) = \\
    \mathbb{E}_{t_s^z, \epsilon_s^z} \Bigg[ \omega(t) \Big( \text{DSSD}_{\text{2D}}^{z}  \Big) \frac{\partial x}{\partial \theta} \Bigg]
\end{multline}
where \( \omega(t) \) is a weighting function regulating timestep contributions.

Further, we optimize the stylized multi-view image \( I_{\text{rgb}} \) and the associated mask \( I_{\text{mask}} \) for alignment with the input data. If required, additional loss terms such as SSIM loss \cite{wang2004image} or LPIPS loss \cite{zhang2018unreasonable} may be integrated to enhance alignment. Thus, our final objective function is:
\begin{equation}
    \mathcal{L}_{\text{style}} = \lambda_{\text{DSSD}} \mathcal{L}_{\text{DSSD}} + \lambda_{\text{RGB}} \mathcal{L}_{\text{RGB}} + \lambda_{\text{mask}} \mathcal{L}_{\text{mask}}
\end{equation}

This setup ensures multi-view consistency in 3D stylization, achieving refined style expression and geometric fidelity through the dynamic coefficient adjustment and adaptive optimization strategy.


\subsection{Low-Level Refinement: Simultaneously Optimized Scale}
To further enhance the texture details of 3D gaussian, multiscale stylization strategy is introduced into the optimization process, called Simultaneously Optimized Scale~(SOS) \cite{galerne2024sgsst}.
Following the silimar approach from \cite{gatys2017controlling}, we employ VGG-19 \cite{simonyan2014very} to extract high-resolution texture features through its shallow convolutional layers. We use N rendered images~(each image is represented as \( I_{\text{v}} \)) from the source 3D gaussian and style reference image \( I_{\text{ref}} \) to compute multi-scale Gram matrix correlations and formulate the style objective as follows:

\begin{equation}
    \mathcal{L}_{\text{SOS}} = \frac{1}{N}\sum_{v=1}^{N}\sum_{l \in L_{s}} \| G(\phi_{\text{VGG}}^l(I_{v})) - G(\phi_{\text{VGG}}^l(I_{\text{ref}})) \|_2^2
\end{equation}

where \( G(\cdot) \) denotes Gram matrix computation,\( \phi_{\text{VGG}}^l \) represents features from the \( l \)-th VGG layer and $L_{s}=\{ReLU\_k\_1, k \in {1, 3, 5}\}$.

\subsection{CLIP-based Style Stream: Contrastive Style Descriptor}

To further align to the style between the 3D gaussian and the given reference image, we utilize Contrastive Style Descriptor~(CSD) \cite{somepalli2024CSD}, which aims to build a high-performance model~(variants of ViT~\cite{dosovitskiy2020image}, like ViT-B and ViT-L) for the representation of the image style. The ViT is trained with both self-supervised learning and supervised objectives, which can extract image descriptors with concise and effective style information. CSD leverage the ViT to extract style feature from rendered images and reference image respectively and then calculate the pairwise cosine silimarity score. Finally, the CSD loss term reduces to:
\begin{equation}
    \mathcal{L}_{\text{CSD}} = \frac{1}{N}\sum^{N}_{v=1}(1 - \cos(\phi_{\text{ViT}}(I_{v}), \phi_{\text{ViT}}(I_{\text{ref}})))
\end{equation}

\subsection{CLIP-based Content Stream: 3D Gaussian Quality Assessment}
\label{sec_qa}
In addition to preserving the original content and migration style of 3D gaussian, we also need to ensure the overall aesthetics quality between the migrated style and content. CLIP-IQA~\cite{wang2022clip_iqa} has been developed to evaluate the look or quality of an image. CLIP-IQA leverages CLIP for perception assessment and calculate the cosine similarity between the feature embeddings of the given text promt and image as follows:
\begin{equation}
    s = \frac{x\odot t}{|| x || * || t ||}
\end{equation}

where $x\in \mathbb{R}^{C}$ and $t\in \mathbb{R}^{C}$ represents the image embedding and text embedding, C is the embedding channel dimension. 
CLIP-IQA further introduces antonym prompts (e.g., “Good photo.” and “Bad photo.”) to address the linguistic ambiguity. $t_{1}$ and $t_{2}$ are obtained from text prompts with good quality  and bad quality respectively and the $s_{i}$ can be obtained with the corresponding $t_{i}$, then the final CLIP-IQA score can be formulate as:
\begin{equation}
     \overline{s} = \frac{e^{s_{1}}}{e^{s_{1}} + e^{s_{2}}}
\end{equation}

We adopt the CLIP-IQA property and extend CLIP-IQA to the 3D style transfer field to ensure the perception quality of 3D gaussian. More specifically, we define the 3D gaussian Quality Assessment~(3DG-QA) as a objective term as:
\begin{equation}
    \mathcal{L}_{\text{3DG-QA}} =  \frac{1}{N}\sum^{N}_{v=1}(1 - \overline{s}_{v})
\end{equation}

where $v$ means the viewpoint index rendered from the 3D gaussian representation.

\subsection{Stylization with Hierarchical Priors}
\label{sec_moe}
\name{} systematically addresses five fundamental aspects in 3D gaussian stylization: (1) Style-content decoupling, (2) Adaptive style conditioning, (3) Multi-scale feature alignment, (4) Texture detail enhancement, and (5) Global aesthetic optimization with four principal components. The DSSD stablishes effective style conditioning through high-level semantic alignment, leveraging score-based stable diffusion to extract and transfer domain-invariant style features. SOS addresses low-level feature alignment via multi-scale optimization, preserving stylistic textures through scale-aware importance sampling and geometric consistency constraints. CSD facilitates mid-level style-content harmonization using contrastive learning to disentangle and recompose style attributes while maintaining content integrity. At last, 3DG-QA enhances global aesthetic quality through metric-guided refinement, employing perceptual quality evaluation to optimize both local textural coherence and global visual appeal.

We integrate the whole optimization goal as:
\begin{multline}
    \mathcal{L}_{final} = \mathcal{L}_{\text{style}}+\mathcal{L}_{\text{SOS}}+\mathcal{L}_{\text{CSD}}+\mathcal{L}_{\text{3DG-QA}} 
\end{multline}

In summary, this multi-faceted approach ensures semantic-aware style, fine-grained style, style fidelity and global aesthetics quality.

As demonstrated in Sec.~\ref{subsec:ablation}, the combined losses enable simultaneous preservation of geometric integrity and artistic expression while suppressing common artifacts like over-stylization and texture flickering.

\section{Experiment}
\label{sec:experiment}
\begin{figure*}[ht]
  \centering
    \includegraphics[width=\linewidth]{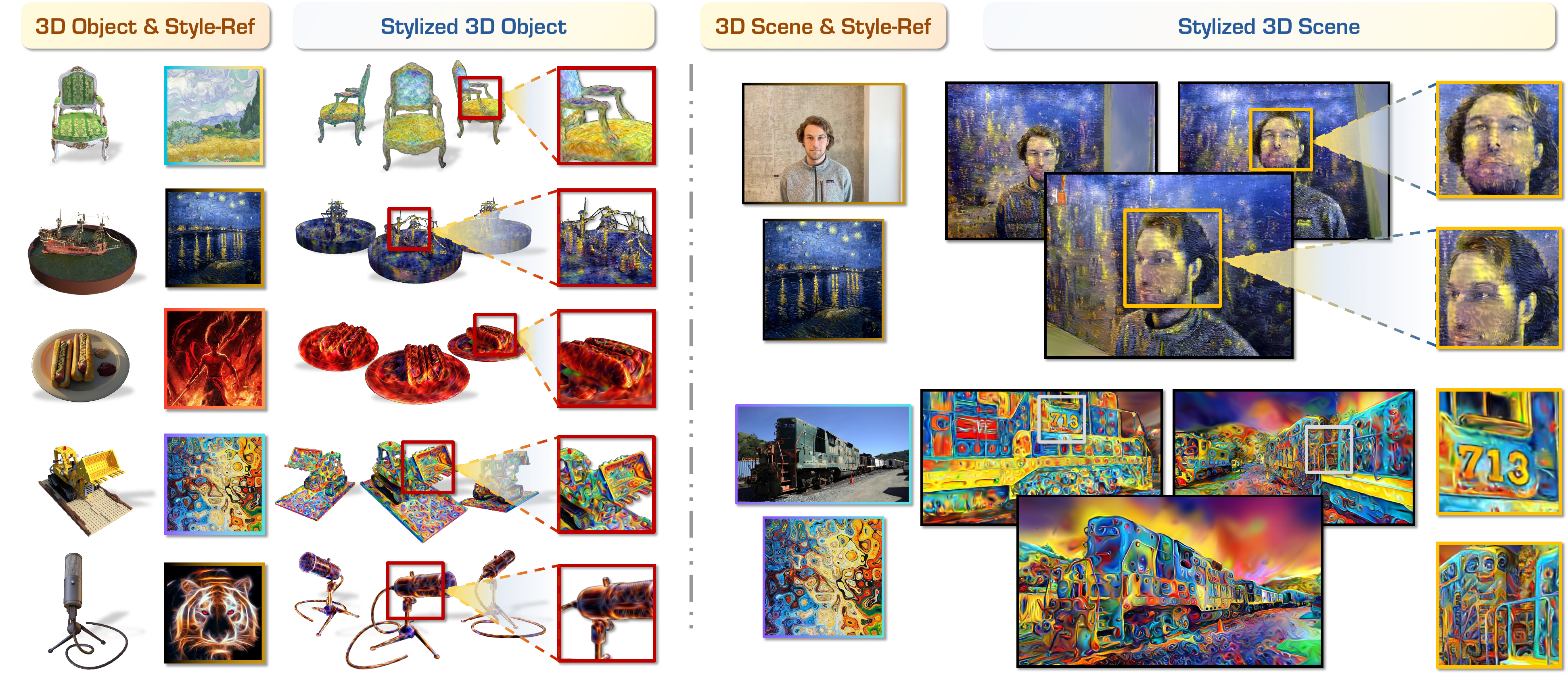}
  \caption{Stylization Visualizations with various reference images. Demonstration of our method's performance across five styles (vangogh wheat field, star night, fire nezha, colorful oil, and lighting tiger) applied to five objects(chair, ship, hotdog, lego and mic) and two scenes (man face and train). The results illustrate our model's capability to handle two main categories of styles: (1) Non-photorealistic Art Styles (\eg cartoon, drawing), showcasing traditional artistic expressions, and (2) State-based Styles (\eg fire, oil), which capture physical properties. This figure demonstrates our method's versatility and semantic-aware ability in stylizing 3D models while preserving style fidelity and geometric consistency across diverse artistic and physical characteristics. For Example, semantic separation of the legs of the chair from the seat cushion, detail texture of chair, texture of the fire on the hot dog, and metallic sheen on the mic are all effectively preserved.}
  \label{fig:visual_result}
\end{figure*}


\subsection{Comparison Studies}

\begin{figure*}[ht]
  \centering
    \includegraphics[width=\linewidth]{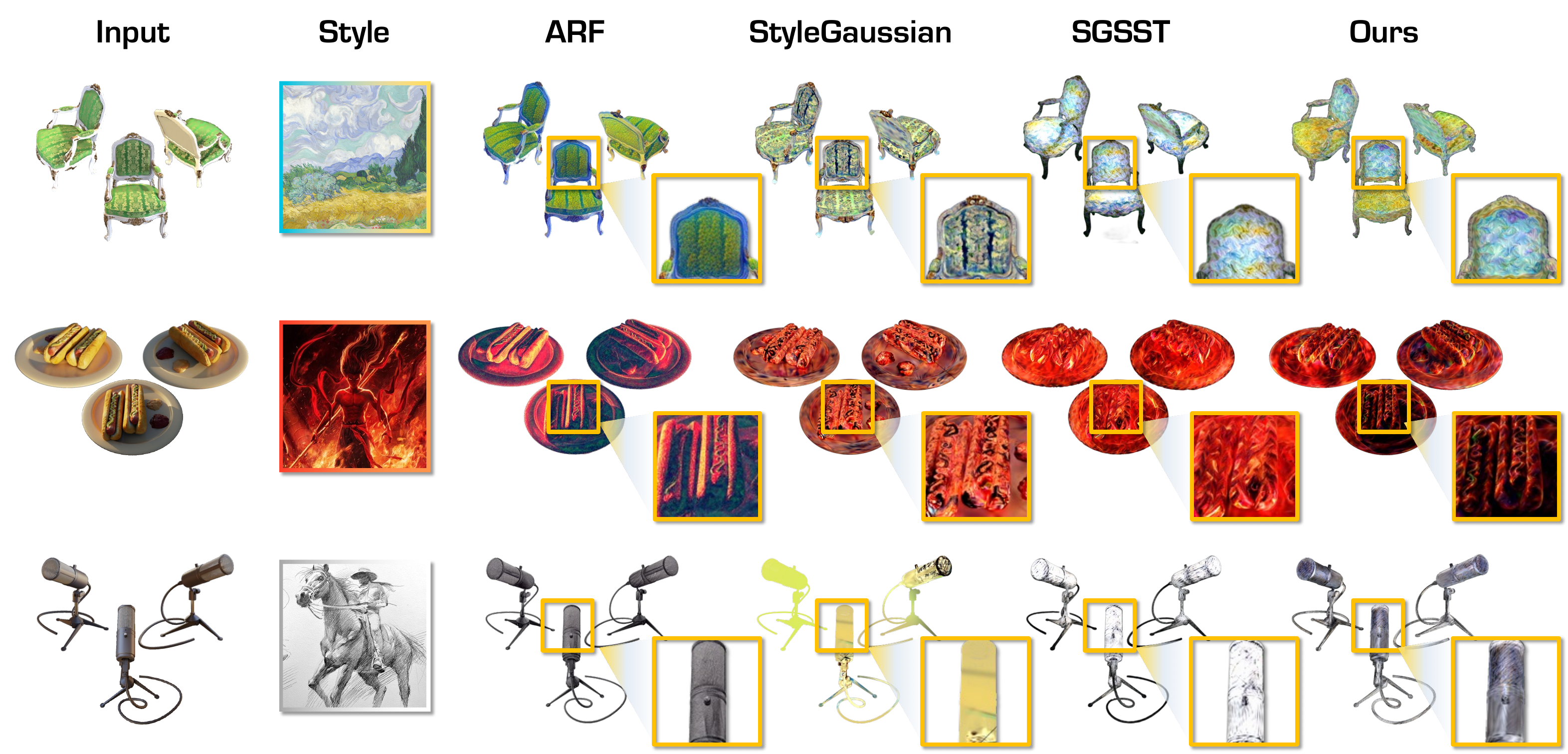}
  \caption{Qualitative Comparisons on Object Level Stylization. We compare our method against other SOTA on nerf synthetic dataset (selected chair, hotdog, and mic) using vangogh wheat field, fire nezha, and sketch styles. The horizontal axis represents the compared methods, and the vertical axis displays different data. Our method effectively retains semantic and details of original model and style feature of reference image, such as semantic separation of the legs of the chair from the seat cushion, texture of the fire on the hot dog, and metallic sheen on the mic. Compared to others, our method exhibits stronger semantic understanding, clearly distinguishing elements like the cushions, backrest and legs on the chair.}
  \label{fig:obj_comparison}
\end{figure*}

\begin{figure*}[!h]
  \centering
    \includegraphics[width=\linewidth]{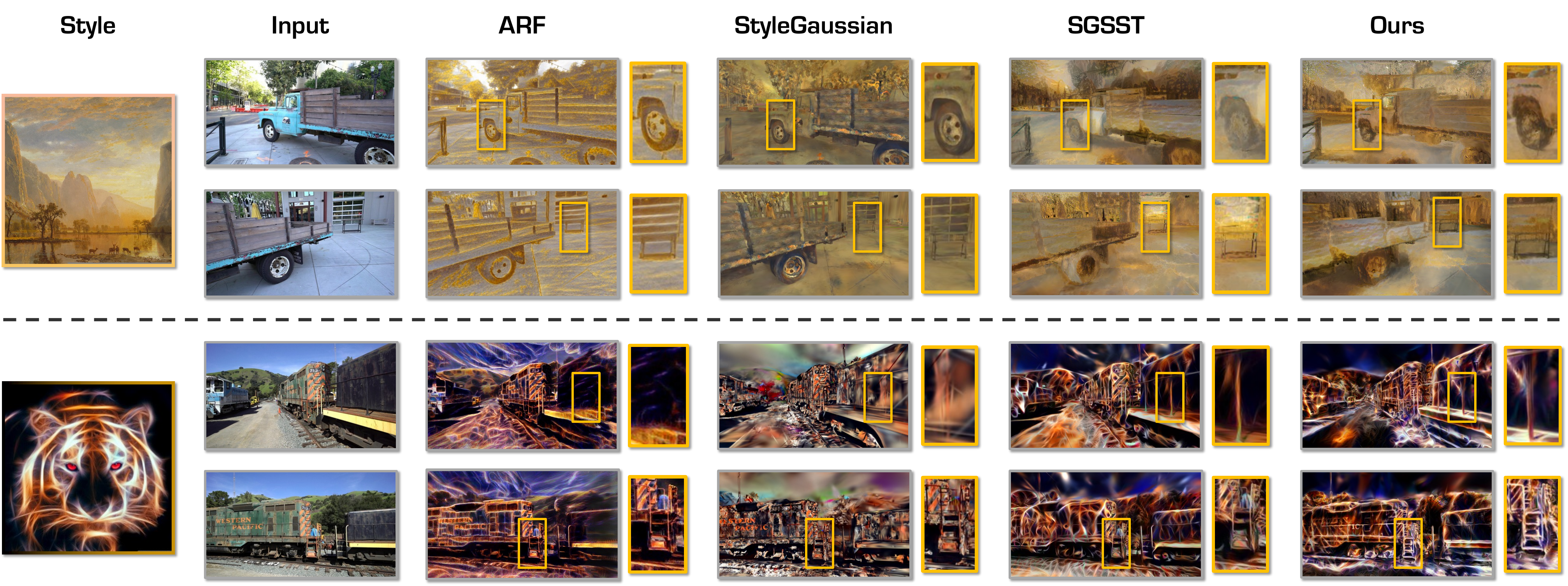}
  \caption{Qualitative Comparisons on Scene Level Stylization. We compare our method against other SOTA on tandt db dataset (selected truck and train) using landscope and lighting tiger styles. The horizontal axis represents the compared methods, and the vertical axis displays different data. Our method effectively retains semantic and details of original model and style feature of reference image, such as the truck wheel and train fence (as shown in Zoom-in). Compared to others, our method exhibits stronger semantic understanding, clearly distinguishing elements like the fence, tire and rail.} 
  \label{fig:scene_comparison}
\end{figure*}

\textbf{Qualitative Result.}
we show objects and scene stylization comparisons in Fig.\ref{fig:obj_comparison} and Fig.\ref{fig:scene_comparison} respectively. For objects, we applied vangogh, fire nezha, and sketch styles to chair, hotdog and mic. For scene stylization, we select truck and train from tandt db dataset  using landscope and lighting tiger styles. We evaluate our method against others, including SGSST \cite{galerne2024sgsst}, StyleGaussian \cite{liu2024stylegaussian} and ARF \cite{zhang2022arf}. The horizontal axis lists competing methods and the vertical axis denotes datasets.

Different from traditional methods based only on VGG networks like SGSST~\cite{galerne2024sgsst}, StyleGaussian~\cite{liu2024stylegaussian} and ARF~\cite{zhang2022arf}, which focus on simple style transfer, our approach prioritizes vivid, expressive and semantic-aware stylization. They relies on VGG networks \cite{simonyan2014very} with empirical-based style decoupling, which limiting style extraction with customized references, our diffusion-based and multi-expert method, pre-trained on large-scale style image-text data, captures style features with greater fidelity. Moreover, training on image-text data enhances semantic understanding, allowing content filtering in CLIP space for precise style extraction.

Unlike ARF \cite{zhang2022arf}, which depends on carefully pre-stylized views for effective color matching and risks texture drift if the initial view is misaligned, our method only requires a single arbitrary style reference image. While we incorporate pre-stylized multi-views, they serve solely for pixel-level style guidance in our outpainting process rather than relying on single-view matching, establishing a distinct way from that of ARF.

\
\\

\begin{table*}[!h]
\centering
\caption{Quantitative comparison with competing methods. CLIP(S) means CLIP-based Style Similarity and CLIP(C) means CLIP-based Content Similarity.}
\label{tab:quantitative_metric}
\begin{tabular}{lccccc}
\toprule
\textbf{Method}         & \textbf{PSNR$\uparrow$} & \textbf{SSIM$\uparrow$} & \textbf{LPIPS$\downarrow$} & \textbf{CLIP(S)$\uparrow$} & \textbf{CLIP(C)$\uparrow$}\\
\midrule
ARF                    & 17.537   & 0.802   & 0.188   & 0.269   & \textbf{0.701} \\
Ref-NPR                & 14.047   & 0.655   & 0.331   & 0.531   & 0.664 \\
SGSST                  & 11.963   & 0.678   & 0.306   & 0.354   & 0.595 \\
StyleGaussian          & 7.279    & 0.129   & 0.558   & 0.320   & 0.610 \\
\textbf{Ours}          & \textbf{18.015} & \textbf{0.830} & \textbf{0.174} & \textbf{0.583} & 0.691 \\
\bottomrule
\end{tabular}
\end{table*}


\noindent\textbf{Quantitative Evaluation}
We evaluate our method with three standard image quality metrics: Peak Signal-to-Noise Ratio (PSNR), Structural Similarity Index Measure (SSIM) \cite{wang2004image}, and Learned Perceptual Image Patch Similarity (LPIPS) \cite{zhang2018unreasonable}. PSNR quantifies pixel-level accuracy, indicating how closely the stylized image matches the original. SSIM measures structural similarity, capturing perceptual features like textures and edges. LPIPS assesses perceptual quality based on deep network features, emphasizing visual similarity as perceived by humans.

As shown in Tab. \ref{tab:quantitative_metric}, our method achieves significantly higher SSIM and PSNR scores, demonstrating enhanced structural and perceptual fidelity compared to SGSST, StyleGaussian and ARF. Our higher PSNR and SSIM score indicates better fidelity in color and texture reproduction while preserving structural details. Furthermore, the LPIPS score, measuring perceptual similarity, supports our method’s superior style consistency and stylization quality across multiple viewpoints.


\subsection{Visualizations of Various Object-Style Pairs}
\hspace*{1em}
As shown in Fig. \ref{fig:visual_result}, we applied six styles to showcase our experimental results on both object and scene datasets~(NeRF synthetic dataset \cite{mildenhall2021nerf} and tandt db~\cite{kerbl20233d}). The style references fall into two main categories: non-photorealistic art styles (\eg vangogh, cartoon, sketch, hand-drawing, watercolor, painting) and state-based styles (\eg fire, water, clouds, hair). These categories highlight our method’s ability to handle traditional art styles and capture realistic physical characteristics in 3D. To highlight our method's advantage in preserving detail textures and shadows, we zoom in on details like the legs and detail texture of the chair, texture of the fire on the hot dog, and metallic sheen on the mic. Experimental results indicate that Gaussian Splatting effectively enhances non-photorealistic and state-based style representations, showing strong adaptability in diverse stylized scenarios. Additional results are provided in the Supplementary.


\begin{figure}[t]
  \centering
   \includegraphics[width=1\linewidth]{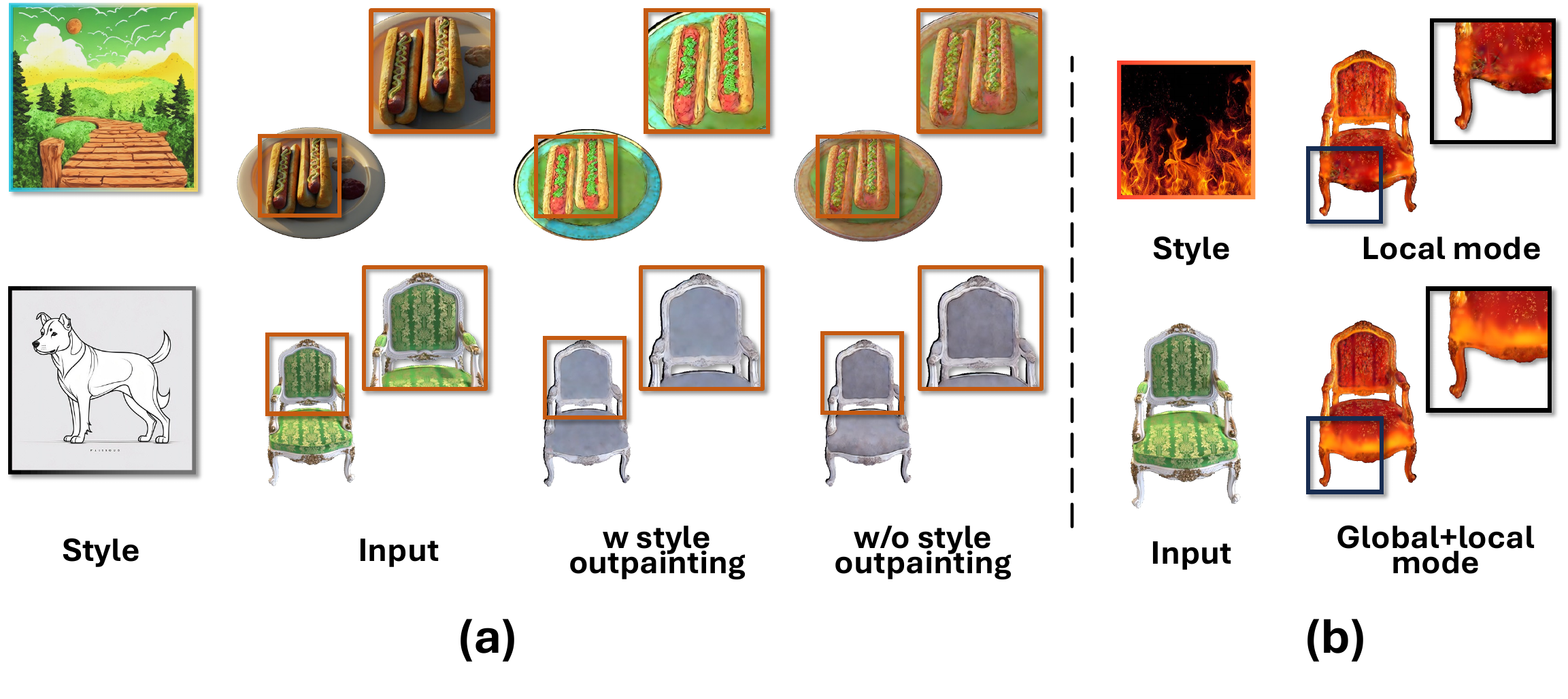}
   \caption{Ablation study on style outpainting guidance mode.
(a) \textbf{Baseline} without style outpainting exhibits limited stylization scope and view-dependent artifacts (red boxes).  
(b) \textbf{Local Guidance} enables single-view enhancement but causes multi-view inconsistencies.  
 \textbf{Global-Local Fusion} achieves cross-view style propagation through adaptive attention weighting, improving style consistency while preserving view-specific details. }
   \label{fig:ablation_outpainting}
\end{figure}

\begin{figure}[t]
  \centering
   \includegraphics[width=1\linewidth]{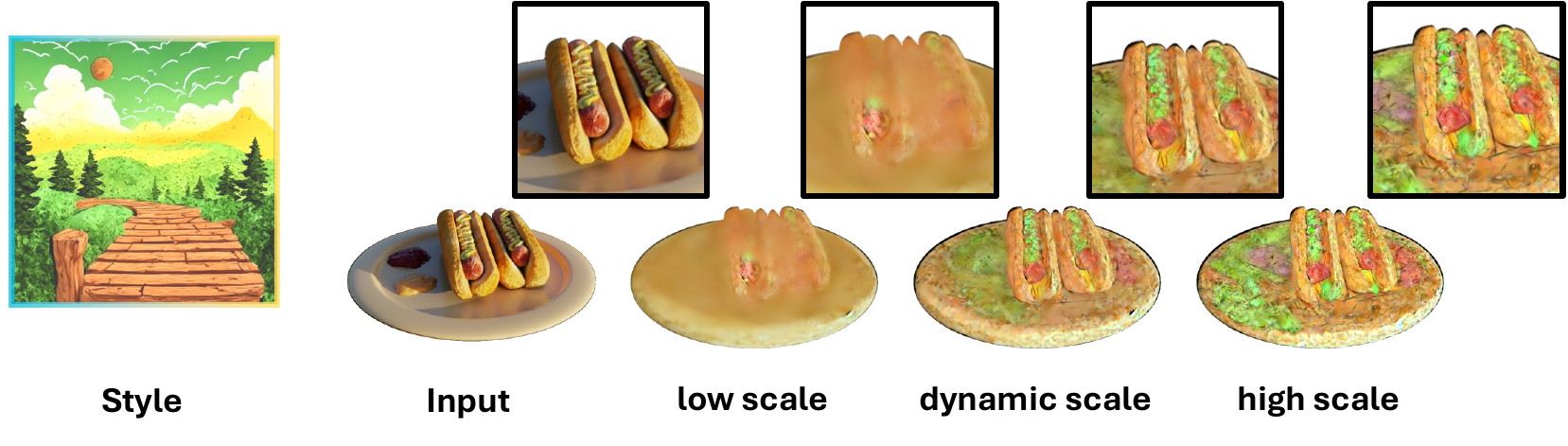}

   \caption{Ablation study on dynamic noise scheduling.
\textbf{Low Scale (7.5)} produces incomplete stylization with missing texture details.  
\textbf{High Scale (50)} introduces oversaturation artifacts and structural distortions.  
\textbf{Dynamic Scale (7.5-30)} adaptively balances detail preservation and style intensity. }
   \label{fig:ablation_scale}
\end{figure}

\begin{figure}[!h]
  \centering
   \includegraphics[width=1\linewidth]{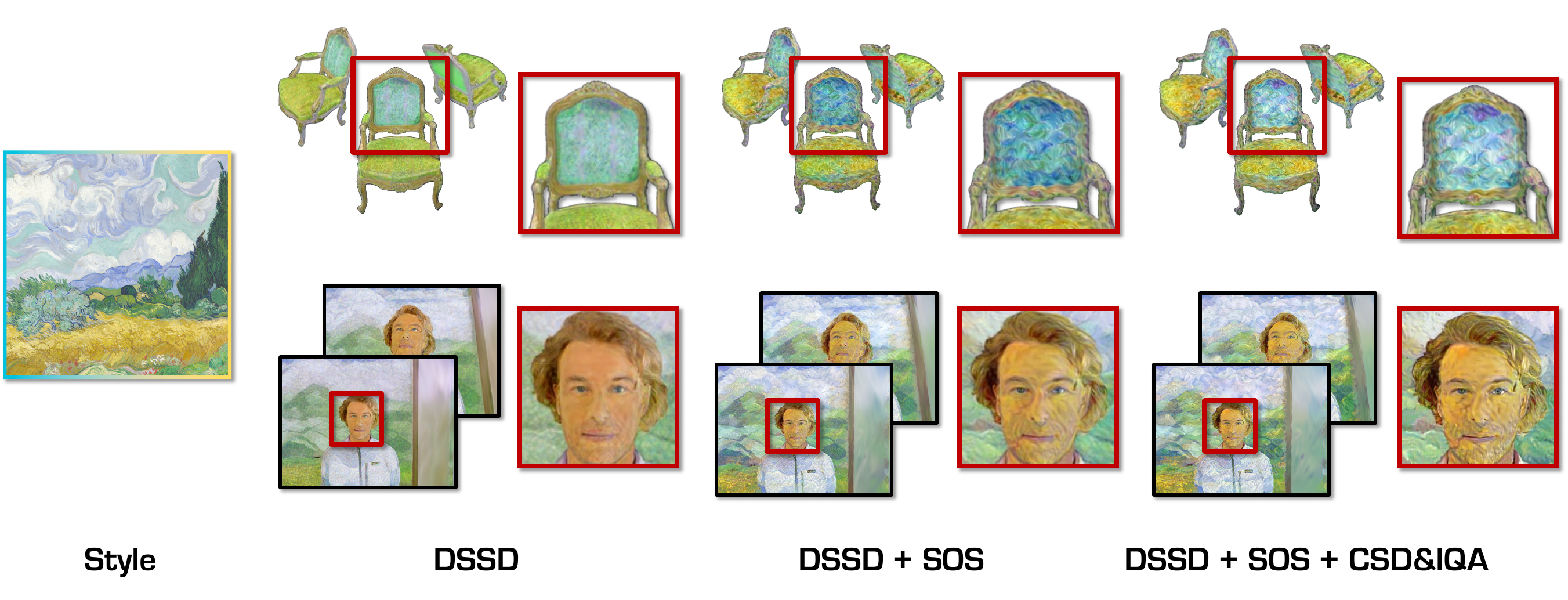}

   \caption{Ablation study loss design.
(a) \textbf{DSSD-only} initialization yields semantically coherent but texture-deficient results with color shifts (see missing curvilinear patterns in Van Gogh stylization).  
(b) \textbf{DSSD+SOS} achieves texture-geometry equilibrium through gradient mutual regularization, recovering fine details while suppressing over-smoothing.  
(c) \textbf{Full Model} enhances perceptual quality via knowledge-driven style assessment, achieving remarkable improvement over baseline (Table~\ref{tab:ablation_quantitative_metric}). }
   \label{fig:ablation_moe}
\end{figure}

\begin{table*}[htp]
\centering
\caption{Quantitative comparison with V1 (DSSD version), V2 (DSSD+SOS version), V3 (DSSD+SOS+CSD version) and V4 (DSSD+SOS+CSD+IQA version). \textbf{V4} is the final Multi-Expert version.}
\label{tab:ablation_quantitative_metric}
\begin{tabular}{lccccc}
\toprule
\textbf{Method} & \textbf{PSNR$\uparrow$} & \textbf{SSIM$\uparrow$} & \textbf{LPIPS$\downarrow$} & \textbf{CLIP(S)$\uparrow$} & \textbf{CLIP(C)$\uparrow$} \\
\midrule
Ours (V1) & 17.270 & 0.776 & 0.181 & 0.280 & \textbf{0.811} \\
Ours (V2) & 17.650 & 0.800 & 0.178 & 0.285 & 0.805 \\
Ours (V3) & 17.900 & 0.820 & 0.175 & 0.325 & 0.695 \\
\textbf{Ours (V4)} & \textbf{18.015} & \textbf{0.830} & \textbf{0.174} & \textbf{0.331} & 0.701 \\
\bottomrule
\end{tabular}
\end{table*}

\subsection{Ablation Study}
\label{subsec:ablation}
\hspace*{1em}
We conducted ablation studies to assess the impact of various components and parameters in our method, focusing on style outpainting mode, DDSD and multi-expert module. 

\noindent\textbf{Ablation on Style Outpainting.}
As shown in Fig. \ref{fig:ablation_outpainting}. We present an ablation study on the impact of Style Outpainting. Without it, the degree of stylization is visibly limited, whereas applying Style Outpainting allows effective style propagation across views. We compares different guidance schemes: local mode \& global-local mode. Local mode shows inconsistencies, resulting in artifacts and missing details in certain views. In contrast, global-local mode enhances stylization intensity and detail refinement, achieving more coherent stylization across views.

\noindent\textbf{Ablation on DSSD.}
As shown in Fig. \ref{fig:ablation_scale}. We conducted an ablation study on the effectiveness of dynamic guidance scale in DSSD. Comparing results at a low scale of 7.5, a high scale of 50, and a dynamic scale ranging from 7.5 to 30, we observed that the dynamic scale approach consistently outperforms static setting.

\noindent\textbf{Ablation on Multi-Expert.} 
As shown in Fig. \ref{fig:ablation_moe},  we analyze the impact of SOS, CSD and 3DG-QA on stylization quality. our analysis reveals that initial stylization using DSSD alone produces semantically coherent results but suffers from two critical limitations: 1) Insufficient low-level texture details (e.g., missing curvilinear patterns in Van Gogh-inspired wheat field renderings), and 2) Systematic color deviation artifacts. The introduction of SOS loss establishes a dual-optimization framework where DSSD and SOS operate concurrently within single-view projections. This configuration enables mutual regularization of their gradient optimization directions - DSSD's tendency toward over-smoothing is counterbalanced by SOS's capacity for detail enhancement, while SOS's potential over-emphasis on low-level features is constrained by DSSD's semantic guidance.

Subsequent integration of CSD and 3DG-QA implements knowledge-driven perceptual assessment through CLIP-space cosine similarity metrics. The CSD module specializes in style authenticity evaluation through learned artistic aesthetics criteria, while 3DG-QA provides quality-focused guidance via antonymic text prompts. Quantitative analysis shows this combined approach achieves remarkable improvement in human perceptual quality scores compared to baseline configurations (see Table \ref{tab:ablation_quantitative_metric}).

\section{Conclusion}
\label{sec:conclusion}
We redefine the 3D Gaussian Splatting (3D GS) stylization task through comprehensive analysis and propose the \name{} framework, establishing a novel paradigm for artistic 3D scene stylization. \name{} enables artistic 3D Gaussian Splatting stylization via Diffusion-guided score distillation (DSSD), CLIP-based dual-stream (style and content) alignment (CSD\&3DG-QA), and multi-scale optimization (SOS). Experiments show superior detail retention, style consistency across various objects and scenes.


\bibliographystyle{ieee_fullname}
\bibliography{bibliography}

\clearpage
\appendix

\noindent
\textbf{\LARGE Appendix}

\setcounter{equation}{0}
\setcounter{figure}{0}
\setcounter{table}{0}
\setcounter{section}{0}
\makeatletter
\renewcommand{\theequation}{S\arabic{equation}}
\renewcommand{\thefigure}{S\arabic{figure}}
\renewcommand{\thetable}{S\arabic{table}}


\section{Preliminary}

\subsection{Style-aware Image Customization}
\hspace*{1em}
In recent advancements in style transfer, StyleShot \coloredcite{cityblue}{gao2024styleshot} and IP-Adapter \coloredcite{cityblue}{ye2023ip} represent two prominent techniques, each employing distinct methods to transfer style from a reference image to a target image. StyleShot emphasizes the extraction of detailed style features using a style-aware encoder, which leverages multi-scale patch partitioning to capture both low-level and high-level style cues. Specifically, StyleShot divides the reference image into non-overlapping patches of three sizes, corresponding to different scales. For each patch scale, there is a dedicated ResBlock at different depths.

The following are the key formulas for style injection in StyleShot:

\begin{equation}
\text{Attention}(Q, K_{s}, V_{s}) = \operatorname{softmax}\left(\frac{Q K_{s}^{T}}{\sqrt{d}}\right) \cdot V_{s}
\end{equation}
where \( Q \) is the query projected from the latent embeddings \( f \), and \( K_{s} \) and \( V_{s} \) are the keys and values, respectively, that the style embeddings \( f_{s} \) are projected onto through independent mapping functions \( W_{K_{s}} \) and \( W_{V_{s}} \). The attention outputs of the text embeddings \( f_{t} \) and style embeddings \( f_{s} \) are then combined into new latent embeddings \( f' \), which are fed into subsequent blocks of Stable Diffusion:

\begin{equation}
f' = \operatorname{Attention}(Q, K_{t}, V_{t}) + \lambda \operatorname{Attention}(Q, K_{s}, V_{s})
\end{equation}
where \( \lambda \) represents the weight balancing the two components.

\subsection{Score Distillation Sampling for 3D Generation}
\hspace*{1em}
Text-guided 3D generation has gained significant attention due to advancements in methods such as Score Distillation Sampling (SDS) \coloredcite{cityblue}{poole2022dreamfusion}, which facilitates the optimization of 3D representations using pre-trained diffusion models. SDS optimizes the parameters $\theta$ of a 3D model $g(\theta)$ by distilling gradients from a diffusion model $\phi$, ensuring that 2D projections generated from $g(\theta)$ align with a target text prompt. The gradient of the SDS loss is defined as:
\begin{equation}
\nabla_\theta L_{\text{SDS}}(\phi, x = g(\theta)) = \mathbb{E}_{t, \epsilon} \left[ \omega(t) \left( \hat{\epsilon}_\phi(z_t; y, t) - \epsilon \right) \frac{\partial x}{\partial \theta} \right],
\end{equation}
where $\hat{\epsilon}_\phi(z_t; y, t)$ represents the predicted noise residual from the pre-trained diffusion model, $\epsilon$ is the actual noise used in the forward process, $z_t$ is the latent variable at timestep $t$, and $\omega(t)$ is a timestep weighting function.

These have been extended to artistic scene generation \coloredcite{cityblue}{li2024art3d} and combined input conditions, including text and images \coloredcite{cityblue}{wang2024phidias, yan2024dreamdissector}. Recent advances leveraging latent diffusion models have improved the scope and expressiveness of text-to-3D synthesis \coloredcite{cityblue}{yan2024dreamdissector, zhou2024diffgs}, supporting more nuanced and creative 3D outputs.

\subsection{3D Gaussian Splatting}
\hspace*{1em}
3D Gaussian Splatting (3D GS) \coloredcite{cityblue}{kerbl20233d} represents a 3D scene using a collection of spatial Gaussians. Each Gaussian \( g_i \) is defined by a mean position \( \mu_i \in \mathbb{R}^3 \) and a covariance matrix \( \Sigma_i \in \mathbb{R}^{3 \times 3} \), which determines its shape and orientation. The Gaussian’s influence on a point \( \mathbf{x} \) is given by:

\begin{equation}
G(\mathbf{x}) = e^{-\frac{1}{2} (\mathbf{x} - \mu_i)^\top \Sigma_i^{-1} (\mathbf{x} - \mu_i)}
\end{equation}
where \( \Sigma_i = \mathbf{R} \mathbf{S} \mathbf{S}^\top \mathbf{R}^\top \) is decomposed into a rotation \( \mathbf{R} \) and scaling \( \mathbf{S} \) matrices. Each Gaussian has an opacity \( \alpha_i \) and a view-dependent color \( c_i \).

During rendering, Gaussians are projected to 2D and blended using alpha compositing. The final pixel color \( C \) is calculated as:

\begin{equation}
C = \sum_{i=1}^n c_i \alpha_i \prod_{j=1}^{i-1} (1 - \alpha_j)
\end{equation}

Here, \( \alpha_i \) is the effective opacity of the \( i \)-th Gaussian in sorted depth order. Gaussian Splatting enables real-time, differentiable rendering and can reconstruct scenes with multi-view supervision.

Compared to NeRF \coloredcite{cityblue}{mildenhall2021nerf}, 3D Gaussian Splatting is significantly more efficient in both time and memory usage. By representing scenes with Gaussian primitives rather than dense neural networks, it allows for faster rendering and lower computational costs, making it more suitable for real-time applications.


\section{Implementation Details}
\textbf{Computational Environment}:  
All experiments were conducted on a single NVIDIA L40S GPU with 46GB of VRAM.
\\
\textbf{Dataset}:  
NeRF synthetic dataset \coloredcite{cityblue}{mildenhall2021nerf} and tandt db \coloredcite{cityblue}{kerbl20233d}, was used for all experiments.

\subsection{Details of Dynamic Style Score Distillation (DSSD)}
\begin{enumerate}
    \item \textbf{Backbone Models}:  
    For the style-aware diffusion model, we adopt StyleShot, which builds on IP-Adapter and incorporates a style-aware encoder to enhance style representation, enabling robust style transfer through score distillation guidance.

    \item \textbf{Fine Timestep Sampling}:  
    We employ a fine-grained timestep sampling strategy with a timestep constant \(\text{T} = 1000\). Minimum and maximum timesteps were set as \(T_{\text{min}} = 0.02 \cdot \text{T}\) and \(T_{\text{max}} = 0.75 \cdot \text{T}\), respectively. The noise intensity was dynamically reduced to high, medium, and low levels to stabilize the updates during training.

    \item \textbf{Dynamic Guidance Coefficients}:  
    The dynamic guidance coefficient \(\Delta \lambda\) was tuned to adapt to varying scales of the dataset and style variations. For the NeRF Synthetic dataset, we selected \(\lambda_{\text{max}} = 20\) and confined \(\Delta \lambda\) within \([7.5, 20]\).

    \item \textbf{Guidance Modes and Outpainting Strategy}:  
    A total of 2800 steps were employed, segmented into specific guidance modes:
    \begin{itemize}
        \item \textbf{Main RGB Loss (Local Mode)}: Steps 100 to 600.
        \item \textbf{Adaptive Iteration (Global Mode)}: Steps 1 to 1000, alternating between global RGB and global SDS losses.
        \item \textbf{Fixed or Free Global Modes}: Steps 1000 to 1900, alternating between global-fix and global-free modes.
        \item \textbf{Local Mode}: Steps 1900 to 2800.
    \end{itemize}
    This hybrid strategy begins with global optimization before transitioning to local refinement, requiring 1800 iterations for SDS loss.

    \item \textbf{Iteration Time and Cost Analysis}:  
    \begin{itemize}
        \item \textbf{Average Time Per Iteration}: Single-view RGB loss averaged 0.1 seconds, while SDS loss averaged 2.5 seconds.
        \item \textbf{Total Iteration Count and Convergence}: Using RGB loss for the initial 1000 steps and SDS loss for the subsequent 2000 steps, convergence was achieved in approximately 2600 seconds. For enhanced local convergence, an additional 500 to 1000 SDS iterations were applied.
    \end{itemize}
\end{enumerate}

\subsection{Details of Simultaneously Optimized Scale (SOS)}
\begin{enumerate}
    \item \textbf{VGG Feature Extraction} 
    \begin{itemize}
        \item Style layers: \\ \texttt{['r11','r21','r31','r41','r51']}
        \item Content layer: \texttt{['r42']}
        \item Gram matrix weights: \texttt{[1e3/64², 1e3/128², 1e3/256², 1e3/512², 1e3/512²]}
    \end{itemize}
    
    \item \textbf{Two-Phase Optimization}
    \begin{itemize}
        \item Pretraining phase:
        \begin{itemize}
            \item Trigger: \texttt{optimize\_iteration=10000} and current\_iter $<$ 10000
            \item Fixed scale: \texttt{optimize\_size=0.5} (uses minimum resize\_images if unspecified)
            \item Downsampling: Bilinear interpolation \texttt{mode="bilinear"}
        \end{itemize}
        \item Full multi-scale phase: Activates all \texttt{resize\_images} scales
    \end{itemize}
    
\end{enumerate}

\subsection{Details of Contrastive Style Descriptor (CSD)}
    \begin{itemize}
        \item Deployed CSD ViT-L style encoder pretrained on LAION-Styles dataset.
    \end{itemize}
    
\subsection{Details of 3D Gaussian Quality Assessment (3DG-QA)}
    \begin{itemize}
        \item Integrated CLIP-ViT-B with antonymic prompts: \textit{"Good, Sharp, Colorful"} vs \textit{"Bad, Blurry, Dull"}, prompts=(\textit{"quality", "sharpness", "colorfullness"})
        \item \texttt{loss = 1 - (0.4*scores['quality'] + 0.4*scores['sharpness'] + 0.2*scores['colorfullness']).mean()}
        , where \texttt{$w_q=0.4$, $w_s=0.4$, $w_c=0.2$} denote quality, sharpness, and colorfulness weights respectively.
    \end{itemize}


\begin{figure}[t]
  \centering
   \includegraphics[width=\linewidth]{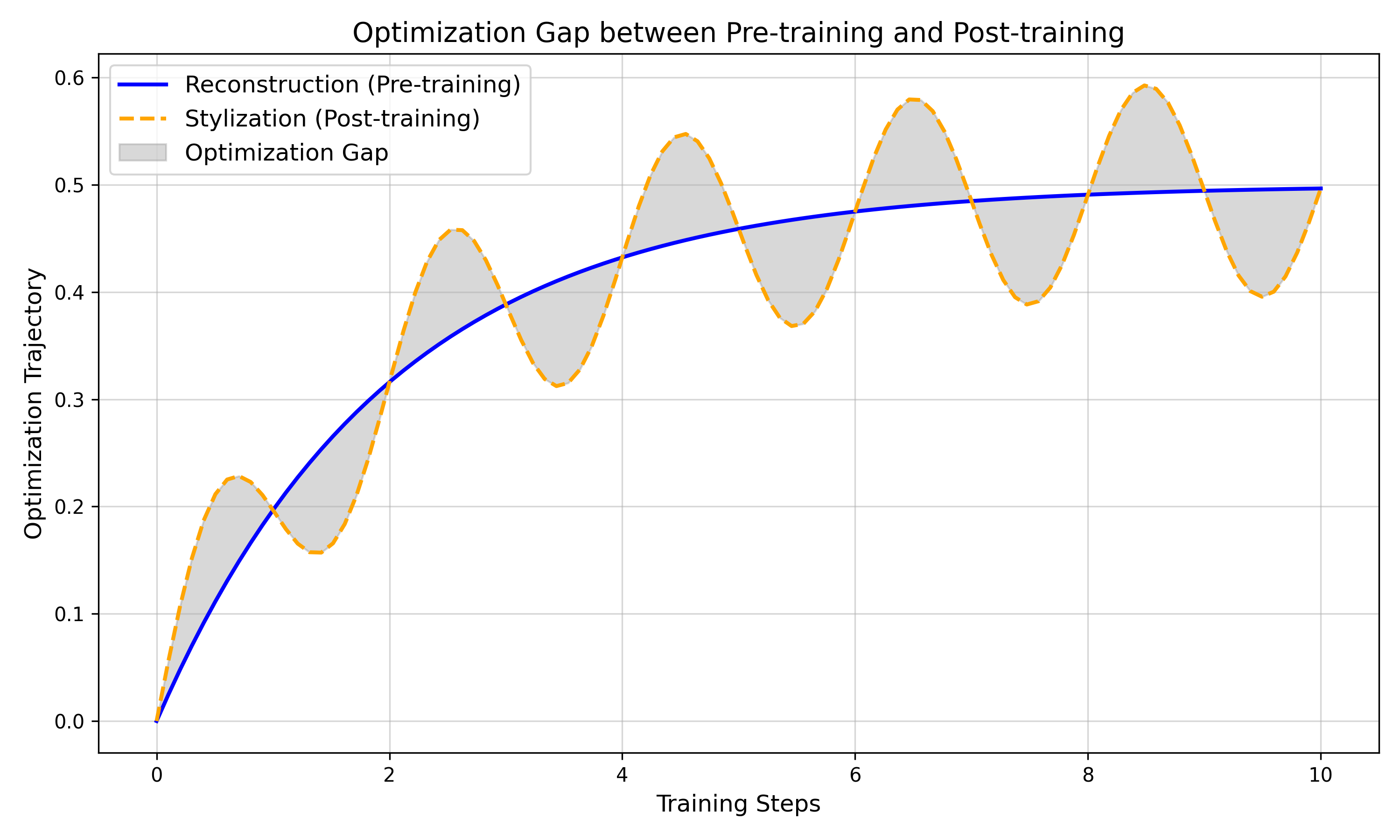}
   \caption{\textbf{Optimization Pathways for Pre-training vs. Post-training.} 
The plot illustrates the optimization pathways for \textit{pre-training} (blue solid line) and \textit{post-training} (orange dashed line), highlighting the \textit{optimization gap} (gray shaded area) between 3D reconstruction and stylization. The pre-training pathway shows smooth, steady convergence, while the post-training pathway oscillates due to inherent uncertainty in stylization. The optimization gap represents misalignment between the stages, emphasizing the need for alignment techniques, such as \textit{style-aware priors} and \textit{dynamic guidance}, to achieve stable and consistent 3D stylization.
}
   \label{fig:opt_gap}
\end{figure}


\section{Additional Method Analysis}
\hspace*{1em}
The challenges of directly transferring 3D generation techniques to 3D stylization stem from the optimization gap between pre-training and post-training stages. This section provides a theoretical and visual analysis of this gap.

\subsection{Misalignment in Optimization Pathways}
\begin{itemize}
    \item \textbf{Pre-training Objective}: The goal of 3D reconstruction during pre-training is to capture geometric and photometric properties accurately. This optimization process is typically smooth and guided by explicit ground truth data.
    \item \textbf{Post-training Objective}: In the post-training phase, the focus shifts to aesthetic alignment using style-aware guidance, which lacks explicit supervision and introduces higher uncertainty.
    \item \textbf{Disjoint Loss Landscapes}: The loss landscapes for pre-training and post-training differ significantly. Pre-training minimizes reconstruction errors, while stylization involves abstract priors from style information, leading to potential misalignment.
\end{itemize}

The optimization pathways during pre-training and post-training can be represented as two distinct loss functions: 
\begin{equation}
    \mathcal{L}_{\text{pre}} = \mathcal{L}_{\text{recon}}(G_{\text{pre}}(x), x_{\text{gt}}),
\end{equation}
where \(\mathcal{L}_{\text{recon}}\) minimizes geometric and photometric errors between the predicted \(G_{\text{pre}}(x)\) and ground truth \(x_{\text{gt}}\), and:
\begin{equation}
    \mathcal{L}_{\text{post}} = \mathcal{L}_{\text{style}}(G_{\text{post}}(x), s_{\text{ref}}),
\end{equation}
where \(\mathcal{L}_{\text{style}}\) aligns the generated results \(G_{\text{post}}(x)\) with a style reference \(s_{\text{ref}}\) using abstract priors.

The optimization gap can then be formulated as:
\begin{equation}
    \Delta \mathcal{L} = \left|\mathcal{L}_{\text{pre}} - \mathcal{L}_{\text{post}}\right|,
\end{equation}
where \(\Delta \mathcal{L}\) quantifies the divergence between the loss landscapes, reflecting the mismatch in optimization objectives.

\subsection{High Uncertainty in Style Information}
\begin{itemize}
    \item \textbf{Multi-modal Style Representations}: Styles are inherently diverse and lack well-defined ground truth, making the optimization process less predictable.
    \item \textbf{Temporal Instability}: Stylization optimization pathways often exhibit oscillations due to conflicts between style priors and geometric constraints.
\end{itemize}

The uncertainty in style optimization can be modeled as the variance in style priors:
\begin{equation}
    \sigma^2_{\text{style}} = \text{Var}(s_{\text{ref}}),
\end{equation}
where \(s_{\text{ref}}\) represents multi-modal style representations. Temporal oscillations in optimization can be expressed as:
\begin{equation}
    \delta_t = \left|\nabla \mathcal{L}_{\text{post}, t+1} - \nabla \mathcal{L}_{\text{post}, t}\right|,
\end{equation}
where \(\delta_t\) measures the instability between consecutive timesteps \(t\) and \(t+1\).

\subsection{Visualization Analysis}
\hspace*{1em}
The graph (Figure \coloredref{citypink}{fig:opt_gap}) visualizes the optimization gap between pre-training and post-training:
\begin{itemize}
    \item \textbf{Pre-training pathway} (blue solid line) shows smooth convergence, reflecting steady optimization for geometric fidelity.
    \item \textbf{Post-training pathway} (orange dashed line) exhibits oscillations, driven by the abstract and subjective nature of style priors.
    \item \textbf{Optimization gap} (gray shaded area) represents the divergence between the two pathways, indicating the challenges of transitioning between the stages.
\end{itemize}

To bridge the optimization gap, alignment strategies must minimize:
\begin{equation}
    \min_{G} \Delta \mathcal{L} + \lambda_{\text{cons}} \mathcal{L}_{\text{consistency}},
\end{equation}
where \(\mathcal{L}_{\text{consistency}}\) enforces multi-view consistency, and \(\lambda_{\text{cons}}\) is a weighting factor to balance consistency with style fidelity.

\subsection{Key Observations and Insights}
\begin{enumerate}

    \item \textbf{Mismatch in Optimization}: 
    The smooth convergence of pre-training contrasts with the oscillatory adjustments in post-training, reflecting the differences in objectives—geometric accuracy vs. subjective style transfer.
    
    The loss landscapes \(\mathcal{L}_{\text{pre}}\) and \(\mathcal{L}_{\text{post}}\) differ fundamentally in their curvature:
    \begin{equation}
        \kappa_{\text{pre}} \ll \kappa_{\text{post}},
    \end{equation}
    where \(\kappa\) represents the curvature, indicating smoother optimization for pre-training compared to post-training.

    \item \textbf{Impact of the Gap}: 
    The optimization gap introduces challenges such as:
    \begin{itemize}
        \item \textbf{Optimization Instability}: Misaligned pathways can lead to instability during post-training.
        \item \textbf{Inconsistent Stylization}: Divergent trajectories may result in geometric distortions or incomplete stylization.
    \end{itemize}
    
    Misaligned pathways can exacerbate:
    \begin{itemize}
        \item Instability: \( \Delta \mathcal{L} \) leads to higher gradients:
        \begin{equation}
            \nabla \mathcal{L}_{\text{post}} \gg \nabla \mathcal{L}_{\text{pre}}.
        \end{equation}
        \item Inconsistency: Variance in style priors \(\sigma^2_{\text{style}}\) introduces inconsistencies in multi-view stylization.
    \end{itemize}

    \item \textbf{Bridging the Gap}: 
    Effective strategies such as \textit{style-aware diffusion priors}, \textit{dynamic style score distillation}, and \textit{progressive style outpainting} are critical to aligning pathways and ensuring robust stylization.
    
    Introducing regularization terms:
    \begin{equation}
        \mathcal{L}_{\text{align}} = \lambda_{\text{prior}} \mathcal{L}_{\text{style}} + \lambda_{\text{geo}} \mathcal{L}_{\text{recon}},
    \end{equation}
    where \(\lambda_{\text{prior}}\) and \(\lambda_{\text{geo}}\) balance style fidelity and geometric preservation, helps align the pathways.
    
\end{enumerate}

\subsection{Conclusion}
This analysis highlights the inherent challenges in aligning pre-training and post-training optimization pathways. The visualization emphasizes the need for dedicated techniques to bridge the gap, ensuring high-fidelity and consistent stylization while maintaining geometric coherence.


\section{More Visual Result}
\hspace*{1em}
As shown in Figure \coloredref{citypink}{fig:more_result_chair}, \coloredref{citypink}{fig:more_result_hotdog}, and \coloredref{citypink}{fig:more_result_mic}, we demonstrate our method's performance across nine distinct styles (sky painting, cartoon, watercolor, fire, cloud, Wukong, drawing, color oil, and sketch) on three datasets (chair, hotdog, and mic).

\begin{figure*}[!h]
  \centering
    \includegraphics[height=0.51\textwidth]{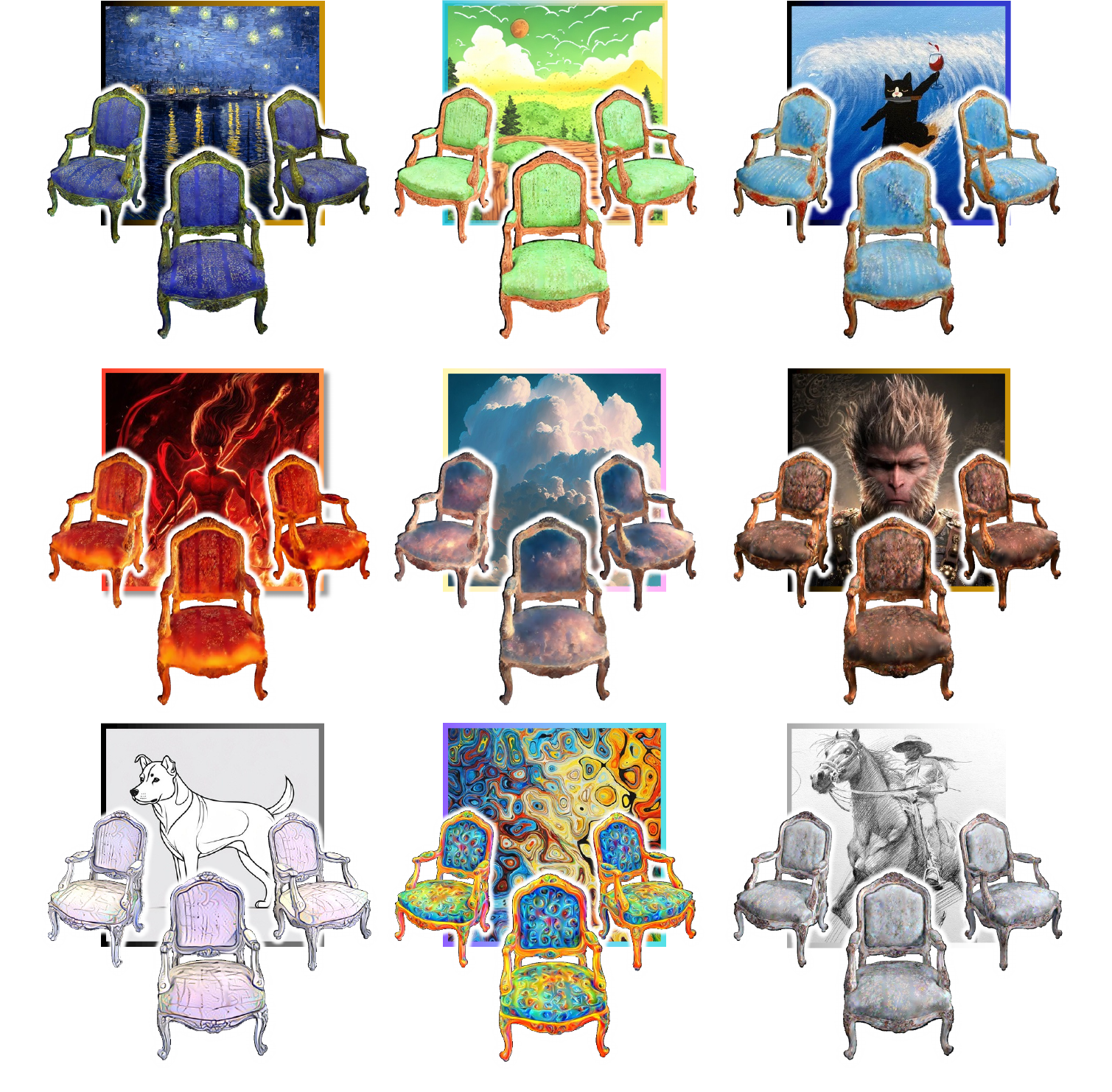}
  \caption{More visual results. Demonstration of our method's performance across nine distinct styles (sky painting, cartoon, watercolour, fire, cloud, Wukong, drawing, color oil, and sketch) applied to chair.}
  \label{fig:more_result_chair}
\end{figure*}

\begin{figure*}[!h]
  \centering
    \includegraphics[height=0.51\textwidth]{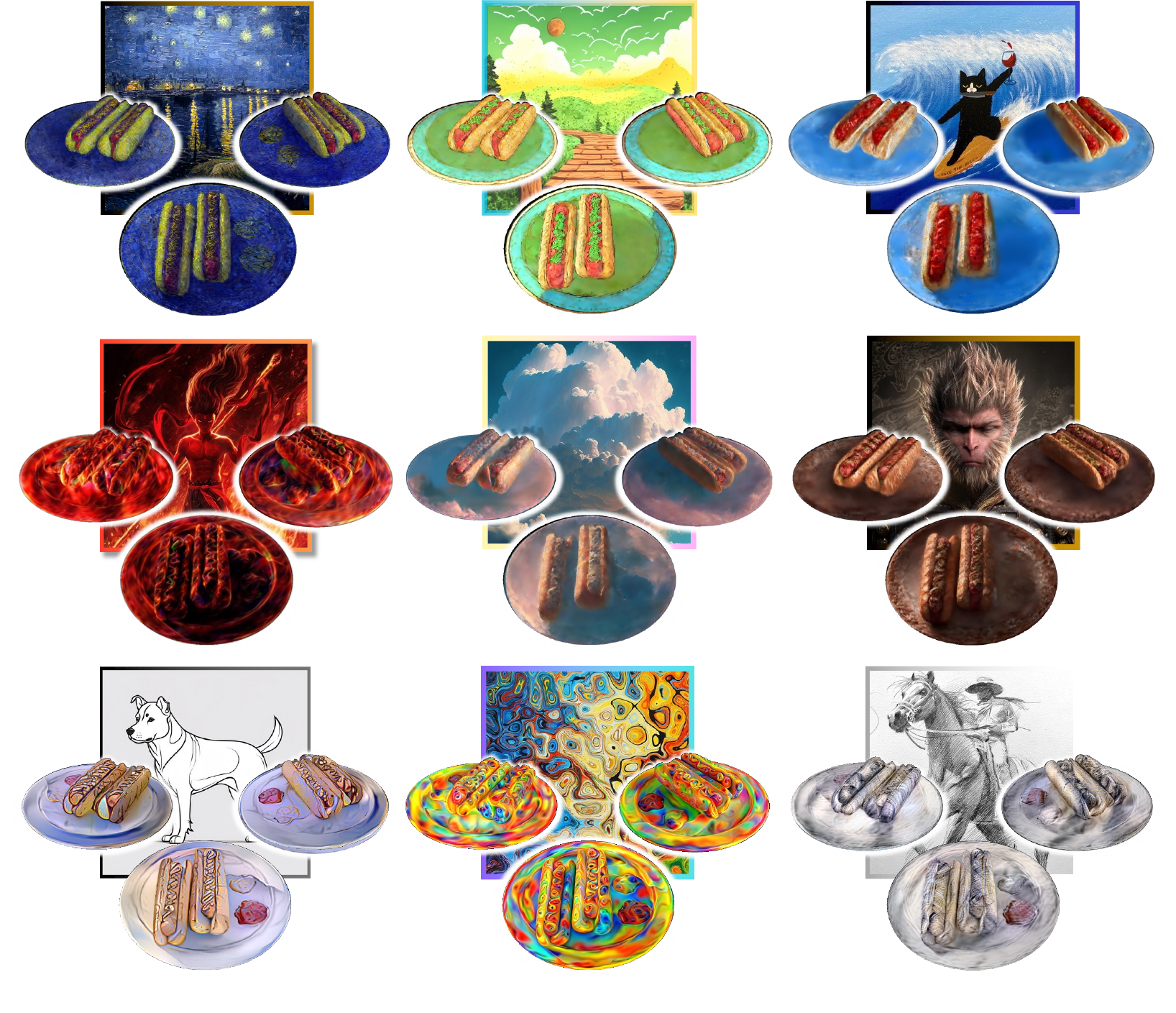}
  \caption{More visual results. Demonstration of our method's performance across nine distinct styles (sky painting, cartoon, watercolour, fire, cloud, Wukong, drawing, color oil, and sketch) applied to hotdog.}
  \label{fig:more_result_hotdog}
\end{figure*}

\begin{figure*}[!h]
  \centering
    \includegraphics[height=0.51\textwidth]{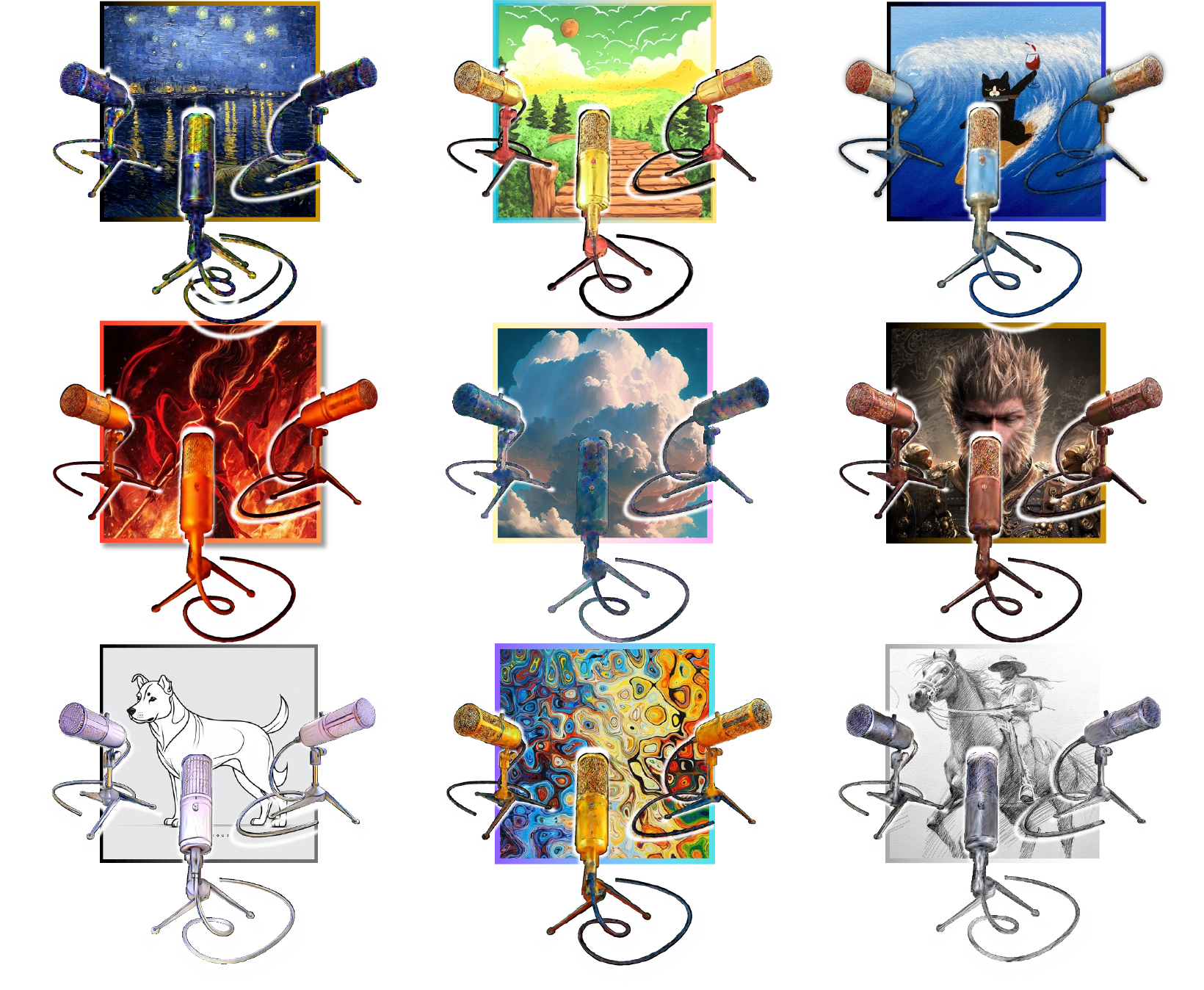}
  \caption{More visual results. Demonstration of our method's performance across nine distinct styles (sky painting, cartoon, watercolour, fire, cloud, Wukong, drawing, color oil, and sketch) applied to mic.}
  \label{fig:more_result_mic}
\end{figure*}

\end{document}